\documentclass[10pt,twocolumn,letterpaper]{article}

\usepackage[pagenumbers]{cvpr} 
\usepackage{graphicx}
\usepackage{amsmath}
\usepackage{amssymb,amsthm}
\usepackage{booktabs}
\usepackage{multirow}
\usepackage{makecell}
\usepackage[dvipsnames,table]{xcolor}
\usepackage{comment}
\usepackage{xspace}
\usepackage{tabularx}
\usepackage[table]{xcolor}
\usepackage{bbding}
\usepackage{pifont}

\usepackage[pagebackref,breaklinks,colorlinks]{hyperref}

\usepackage[capitalize]{cleveref}
\crefname{section}{Sec.}{Secs.}
\Crefname{section}{Section}{Sections}
\Crefname{table}{Table}{Tables}
\crefname{table}{Tab.}{Tabs.}

\def\cvprPaperID{4709} 
\def\confName{CVPR}
\def\confYear{2023}

\begin{document}

\setlength{\abovedisplayskip}{3pt}
\setlength{\belowdisplayskip}{3pt}

\title{Panoptic Lifting for 3D Scene Understanding with Neural Fields}

\makeatother
\newcommand{\MATTHIAS}[1]{{\emph{\textcolor{red}{\textbf{Matthias:~#1}}}}}
\newcommand{\ANGIE}[1]{{\emph{\textcolor{blue}{Angie: #1}}}}
\newcommand{\LORENZO}[1]{{\emph{\textcolor{magenta}{Lorenzo: #1}}}}
\newcommand{\YAWAR}[1]{{\emph{\textcolor{ForestGreen}{Yawar:~#1}}}}
\newcommand{\SAMUEL}[1]{{\emph{\textcolor{brown}{Samuel:~#1}}}}
\newcommand{\PETER}[1]{{\emph{\textcolor{brown}{Peter:~#1}}}}
\newcommand{\TODO}[1]{{\emph{\textcolor{red}{TODO: #1}}}}

\newcommand{\OURS}{Panoptic Lifting\xspace}

\newcommand{\argmax}{\operatornamewithlimits{argmax}}
\newcommand{\argmin}{\operatornamewithlimits{argmin}}
\newcommand{\denselist}{\itemsep 0pt\parsep=0pt\partopsep 0pt\vspace{-\topsep}}

\newcommand{\mypara}[1]{\vspace{4pt}\noindent\textbf{#1}}

\author{
Yawar Siddiqui$^{1,2}$~~~ 
Lorenzo Porzi$^2$~~~
Samuel Rota Bul\`{o}$^2$~~~\\
Norman M{\"u}ller$^{1,2}$~~~
Matthias Nie{\ss}ner$^1$~~~
Angela Dai$^1$~~~
Peter Kontschieder$^2$~~~
\vspace{0.2cm} \\
Technical University of Munich$^1$~~~
Meta Reality Labs Zurich$^2$
\vspace{0.2cm}
}

%%%%%%%%% TEASER

\twocolumn[{%
	\renewcommand\twocolumn[1][]{#1}%
	\vspace{-30pt}
        \maketitle
	\begin{center}
            \vspace{-6mm}
		\includegraphics[width=\linewidth]{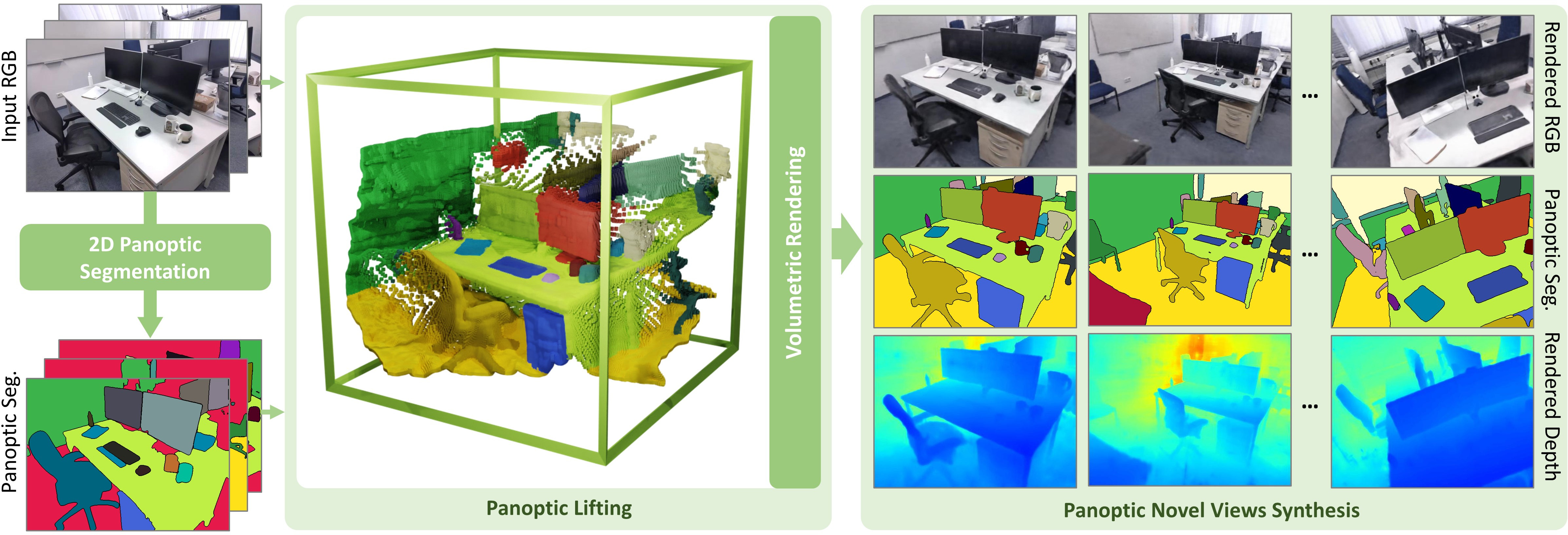}
            \vspace{-6mm}
		\captionof{figure}{Given only RGB images of an \textit{in-the-wild} scene as input, our method optimizes a panoptic radiance field which can be queried for color, depth, semantics, and instances for any point in space. We obtain poses for input images with COLMAP~\cite{schonberger2016structure}, and 2D panoptic segmentation masks using a pretrained off-the-shelf network~\cite{cheng2022masked}. During training, our method lifts these 2D segmentation masks, which are often noisy and view-inconsistent, into a consistent 3D panoptic radiance field. Once trained, our model is able to render images and their corresponding panoptic segmentation masks from both existing and novel viewpoints.}
		
		\label{fig:teaser}
		\vspace{-1mm}
	\end{center}    
}]

\maketitle

%%%%%%%%% ABSTRACT

\begin{abstract}
\vspace{-12pt}\newcommand\blfootnote[1]{%
  \begingroup
  \renewcommand\thefootnote{}\footnote{#1}%
  \addtocounter{footnote}{-1}%
  \endgroup
}

We propose \OURS, a novel approach for learning panoptic 3D volumetric representations from images of in-the-wild scenes.
Once trained, our model can render color images together with 3D-consistent panoptic segmentation from novel viewpoints.
Unlike existing approaches which use 3D input directly or indirectly, our method requires only machine-generated 2D panoptic segmentation masks inferred from a pre-trained network.
Our core contribution is a panoptic lifting scheme based on a neural field representation that generates a unified and multi-view consistent, 3D panoptic representation of the scene.
To account for inconsistencies of 2D instance identifiers across views, we solve
a linear assignment with a cost based on the model's current predictions and the machine-generated segmentation masks, thus enabling us to lift 2D instances to 3D in a consistent way. 
We further propose and ablate contributions that make our method more robust to noisy, machine-generated labels, including test-time augmentations for confidence estimates, segment consistency loss, bounded segmentation fields, and gradient stopping.
Experimental results validate our approach on the challenging Hypersim, Replica, and ScanNet datasets, improving by 8.4, 13.8, and 10.6\% in scene-level PQ over state of the art.
\blfootnote{Project page: \url{nihalsid.github.io/panoptic-lifting/}}

\end{abstract}

%%%%%%%%% BODY TEXT
%-------------------------------------------------------------------------
\vspace{-0.8cm}
\section{Introduction}
\label{sec:intro}
Robust panoptic 3D scene understanding models are key to enabling applications such as VR, robot navigation, or self-driving, and more.
Panoptic image understanding -- the task of segmenting a 2D image into categorical ``stuff'' areas and individual ``thing'' instances -- has experienced tremendous progress over the past years.
These advances can be attributed to continuously improved model architectures and the availability of large-scale labeled 2D datasets, leading to state-of-the-art 2D panoptic segmentation models~\cite{cheng2022masked,yu2022kMeans,li2022mask} that generalize well to unseen images captured in the wild.

\begin{figure}
\begin{center}
\includegraphics[width=0.95\linewidth]{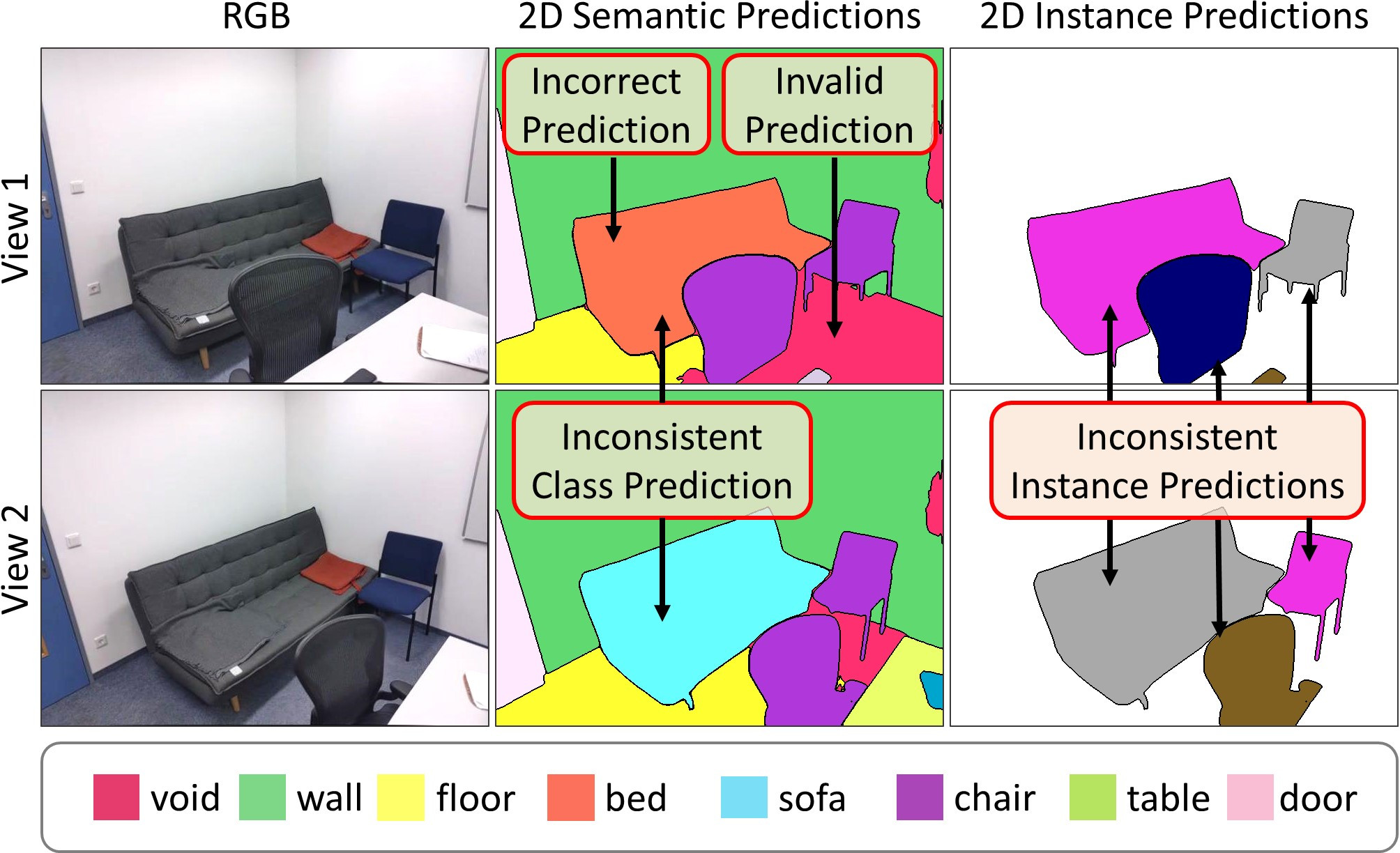}
\vspace{-1mm}
\caption{%
Predictions from state-of-the-art 2D panoptic segmentation methods such as Mask2Former~\cite{cheng2022masked} are typically noisy and inconsistent when compared across views of the same scene.
Typical failure modes include conflicting labels (e.g. sofa predicted as a bed above) and segmentations (e.g. labeled void above).
Furthermore, instance identities are not preserved across frames (represented as different colors).
}
% LORENZO: Changing this to focus more on cross-view errors, otherwise we contradict one of our main motivations for relying purely on 2D panoptic
%Predictions from state-of-the-art 2D panoptic segmentation methods such as Mask2Former~\cite{cheng2022masked} are typically noisy and inconsistent. Within a a single image, semantic segmentations might be incorrect (e.g., sofa predicted as a bed above) or predicted as invalid (shown in red). Across multiple images, the semantics can be inconsistent, i.e., assigning the same point in space conflicting labels. Furthermore, instance labels across the frames are not meaningful, as the instance counting per-frame (shown on the right).}
\label{fig:m2f_inconsistencies}
\end{center}
\vspace{-5mm}
\end{figure}

Single-image panoptic segmentation, unfortunately, is still insufficient for tasks requiring coherency and consistency across multiple views.
In fact, panoptic masks often contain view-specific imperfections and inconsistent classifications, and single-image 2D models naturally lack the ability to track unique object identities across views (see Fig.~\ref{fig:m2f_inconsistencies}).
Ideally, such consistency would stem from a full, 3D understanding of the environment, but lifting machine-generated 2D panoptic segmentations into a coherent 3D panoptic scene representation remains a challenging task.

Recent works~\cite{kundu2022panoptic,zhi2021place,fu2022panoptic,wang2022dm} have addressed panoptic 3D scene understanding from 2D images by leveraging Neural Radiance Fields (NeRFs)~\cite{mildenhall2021nerf}, gathering semantic scene data from multiple sources.
Some  works~\cite{fu2022panoptic,wang2022dm} rely on ground truth 2D and 3D labels, which are expensive and time-consuming to acquire.
The work of Kundu~\etal~\cite{kundu2022panoptic} instead exploits machine-generated 3D bounding box detection and tracking together with 2D semantic segmentation, both computed using off-the-shelf models.
However, this approach is limited by the fact that 3D detection models, when compared to 2D panoptic segmentation ones, struggle to generalize beyond the data they were trained on.
This is in large part due to the large difference in scale between 2D and 3D training datasets.
Furthermore, dependence on multiple pre-trained models increases complexity and introduces potentially compounding sources of error.

In this work we introduce \OURS, a novel formulation which represents a static 3D scene as a panoptic radiance field (see Sec.~\ref{sec:scene_repr}).
\OURS supports applications like novel panoptic view synthesis and scene editing, while maintaining robustness to a variety of diverse input data. 
Our model is trained from only 2D posed images and corresponding, machine-generated panoptic segmentation masks, and can render color, depth, semantics, and 3D-consistent instance information for novel views of the scene.

Starting from a TensoRF~\cite{Chen2022ECCV} architecture that encodes density and view-dependent color information, we introduce lightweight output heads for learning semantic and instance fields.
The semantic field, represented as a small MLP, is directly supervised with the machine-generated 2D labels.
An additional segment consistency loss guides this supervision to avoid optima that fragment objects in the presence of label inconsistencies.
The 3D instance field is modelled by a separate MLP, holding a fixed number of class-agnostic, 3D-consistent surrogate object identifiers.
Losses for both the fields are weighted by confidence estimates obtained by test-time augmentation on the 2D panoptic segmentation model.
Finally, we discuss specific techniques, e.g. bounded segmentation fields and stopping semantics-to-geometry gradients (see Sec.~\ref{sec:loss_fn}), to further limit inconsistent segmentations.

\noindent In summary, our contributions are:
\begin{itemize}
    \item A novel approach to panoptic radiance field representation that models the radiance, semantic class and instance id for each point in the space for a scene by directly lifting machine-generated 2D panoptic labels.
    \item A robust formulation to handle inherent noise and inconsistencies in machine-generated labels, resulting in a clean, coherent and view-consistent panoptic segmentations from novel views, across diverse data.
\end{itemize}

%-------------------------------------------------------------------------
\section{Related Work}
\label{sec:related}
\mypara{Neural Radiance Fields (NeRFs)} offer a unified representation to model a scene's photo-realistic appearance~\cite{barron2021mip,verbin2022ref,Chen2022ECCV,fridovich2022plenoxels,mildenhall2021nerf,mueller2022autorf},  geometry~\cite{wang2021neus, oechsle2021unisurf, Azinovic_2022_CVPR}, and other spatially-varying properties~\cite{zhi2021place,kundu2022panoptic,tschernezki2022neural,vora2021nesf,Gafni_2021_CVPR} (e.g., semantics).
NeRF methods can be broadly divided into two macro-categories: i) those which encode the entire scene into a coordinate-based neural network~\cite{mildenhall2021nerf,barron2021mip,verbin2022ref} and ii) those which attach parameters to an explicit 3D structure, such as a voxel grid~\cite{fridovich2022plenoxels,SunSC22}, point cloud~\cite{xu2022point} or spatial hash~\cite{muller2022instant}.
A speed-memory trade off exists between the two, as models in (i) are generally more compact, while models in (ii) are often faster for training and inference.
In our work we adopt a hybrid approach, modeling appearance and geometry with an explicit neural field derived from TensoRF~\cite{Chen2022ECCV}, and semantics and instances  with a pair of small implicit MLPs.

\mypara{NeRFs for semantic 3D scene modeling.}
The work of Zhi~\etal~\cite{zhi2021place} first explored encoding semantics into a NeRF, demonstrating how noisy 2D semantic segmentations can be fused into a consistent volumetric model, improving their accuracy and enabling novel view synthesis of semantic masks.
Since then, several works have extended this idea, e.g., by adding instance modeling~\cite{kundu2022panoptic,fu2022panoptic,wang2022dm} or by encoding abstract visual features~\cite{tschernezki2022neural, kobayashi2022distilledfeaturefields} from which a semantic segmentation can be derived a-posteriori.
Panoptic NeRF~\cite{fu2022panoptic} and DM-NeRF~\cite{wang2022dm} describe panoptic radiance fields, focusing on the tasks of label transfer and scene editing, respectively.
In contrast to our work, they both require some form of manual ground truth for the target scene: panoptically segmented coarse meshes for the former, and per-image 2D panoptic segmentations for the latter.
On the other hand, Panoptic Neural Fields (PNF)~\cite{kundu2022panoptic} relies purely on RGB images, the same setting as ours.
PNF, however, exploits a much larger set of predictions from pre-trained networks, including per-image semantic segmentation, 3D object bounding boxes, and object tracking across frames.
While 3D tracking enables PNF to handle moving objects, the use of 3D detectors induces strong sensitivity to errors in the predicted boxes, especially beyond the datasets on which these networks were trained on.

\mypara{2D and 3D panoptic segmentation.}
The task of 2D panoptic segmentation and its associated metrics were first defined by Kirillov~\etal~\cite{kirillov2019panoptic}.
A first wave of works in this field~\cite{cheng2020panoptic,kirillov2019panoptic,porzi2019seamless} proposed solutions based on combining a semantic segmentation network with an instance segmentation network, often sharing a common backbone.
Recent works~\cite{cheng2022masked,cheng2021per,zhang2021k} instead adopt a more unified approach, inspired by the DETR object detector~\cite{carion2020end}.
These works use a single transformer-based network to directly produce panoptic output as a set of image segments for both ``things'' and ``stuff.''
In our work, we adopt Mask2Former~\cite{cheng2022masked} with a Swin-L~\cite{liu2021swin} backbone as our pre-trained 2D panoptic segmenter, due to its state-of-the-art performance on the general-purpose COCO Panoptic~\cite{lin2014microsoft} dataset.

Panoptic segmentation has also been explored in a 3D context, both as segmentation of pre-computed 3D structures~\cite{milioto2020lidar,gasperini2021panoster,sirohi2021efficientlps,zhou2021panoptic} (e.g., voxel grids or point clouds), and as simultaneous 3D segmentation and reconstruction from 2D images~\cite{narita2019panopticfusion,rosinol20203d,dahnert2021panoptic}.
However, these methods leveraging meshes, voxel grids or point clouds, do not  allow for photo-realistic novel view synthesis as NeRF-based methods do.

%-------------------------------------------------------------------------

\section{Method}
\label{sec:main}
\begin{figure*}[htp]
\includegraphics[width=\linewidth]{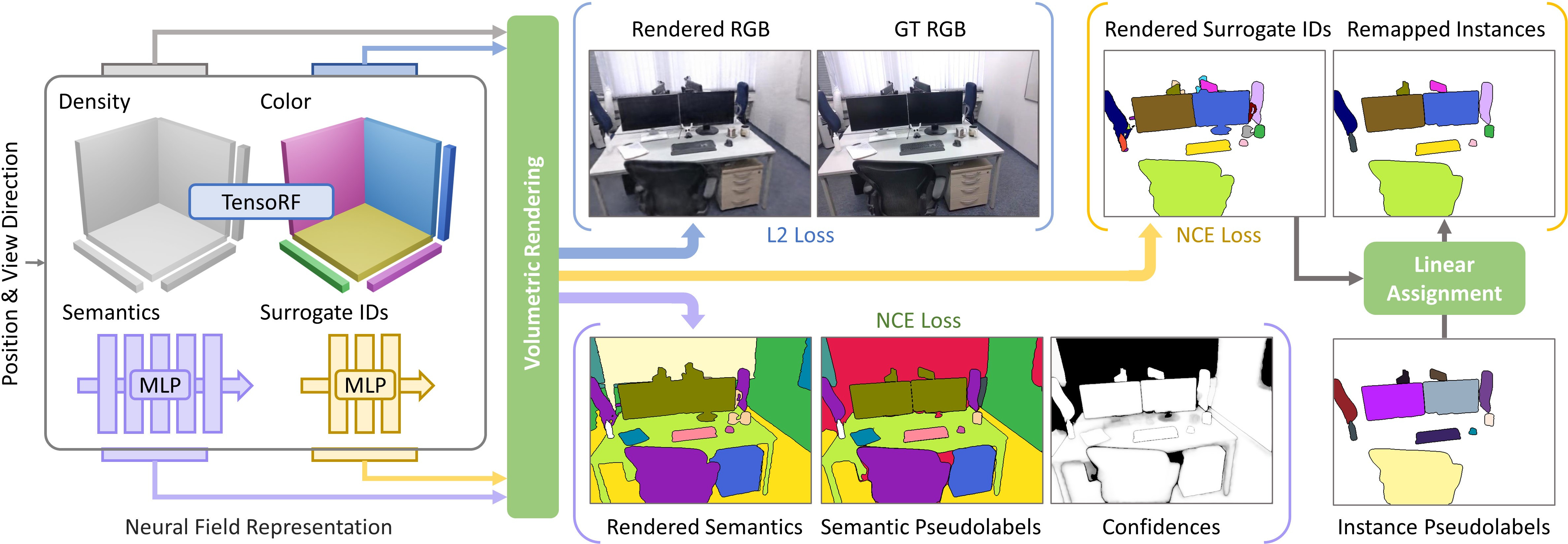}
\vspace{-6mm}
\caption{Our neural field representation comprises color, density, semantics, and instances.
Color and density leverage TensoRF~\cite{Chen2022ECCV}, and we use two small MLPs for semantics and instances. The radiance field is supervised by an L2 photometric loss on the volumetrically rendered radiance along rays. The semantic field is likewise supervised by an NCE loss on the volumetrically rendered class probability distribution and the 2D machine-generated semantic labels. 
%An additional class-agnostic segmentation loss (not shown above) penalizes semantic segmentations if they differ within a 2D machine generated instance. 
For instances, we first map a machine-generated instance to a 3D surrogate identifier using linear assignment, where the cost depends on the current 3D instance predictions. These mapped 2D instances are used to supervise the rendered instance field with an NCE loss. All segmentation losses are weighted by test-time augmentation confidences.}
\label{fig:experiments_scannet}
\vspace{-3mm}
\end{figure*}

\begin{figure}[htp]
\begin{center}
\includegraphics[width=0.95\linewidth]{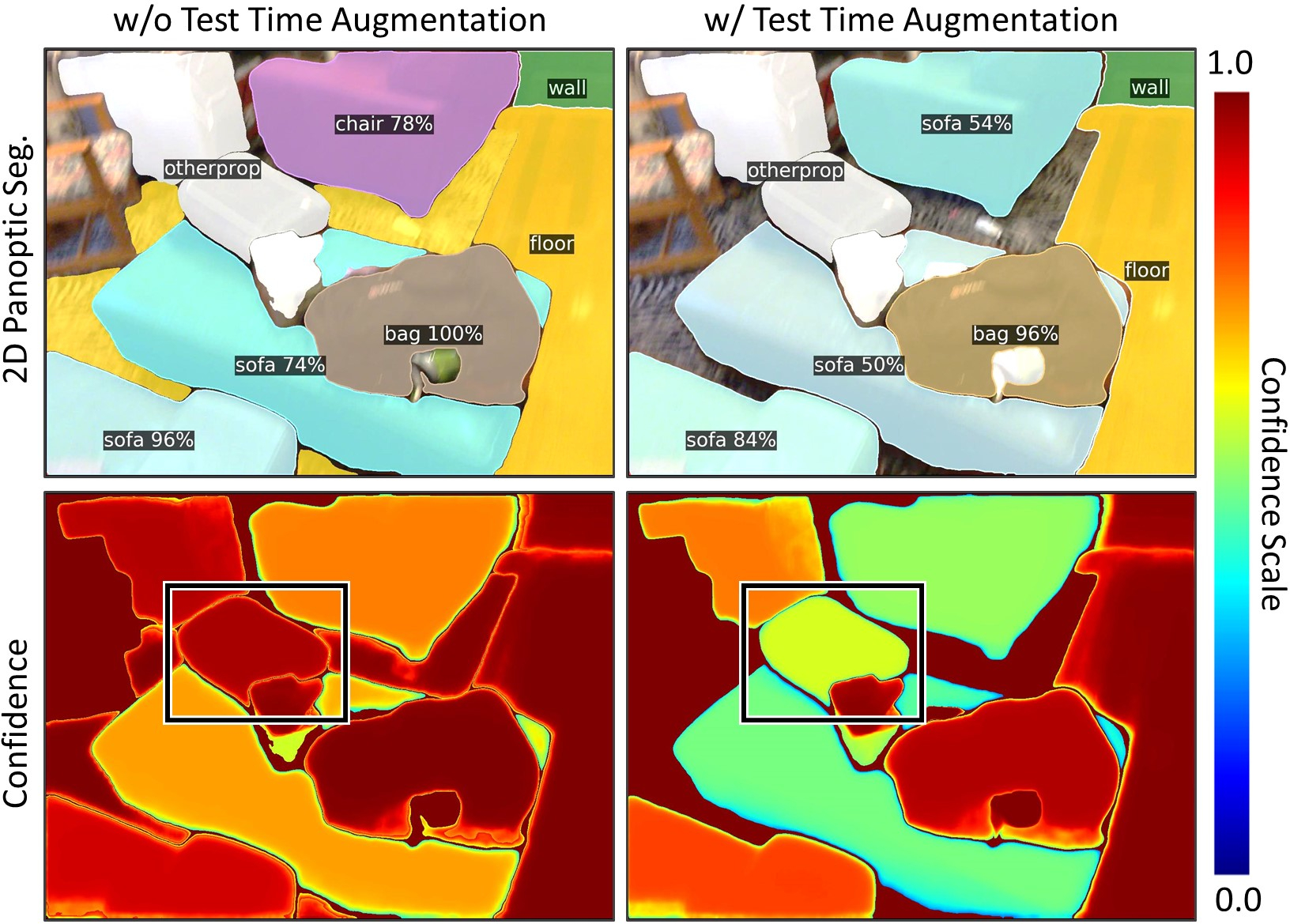}
\vspace{-2mm}
\caption{Confidences from a single Mask2Former~\cite{cheng2022masked} prediction  (left) are often spuriously high, even for incorrectly labeled segments. Above, the \textit{bag} is labeled incorrectly as \textit{otherprop} with high confidence. Merging segmentations from test-time augmentations results in more reliable confidence estimates (right).}
\label{fig:tta_confidences}
\vspace{-7mm}
\end{center}
\end{figure}

Given posed RGB images $\{I\}$ of a scene with corresponding machine-generated 2D panoptic segmentations, our goal is to build an implicit, volumetric representation of the scene that models appearance and density, together with 3D object instances and their semantics.
Our approach, called \OURS, enables generating 2D panoptic segmentations from novel views in a 3D-consistent fashion, such that the rendered 2D instance ID of the same 3D object is preserved across views.

\subsection{Input Data}
\label{sec:input_data}
Input posed images $\{I\}$ are abstracted as sets of viewing rays expressed in world coordinates. 
A ray $r$ can be parameterized in 3D space as $r=p_0+td_r$, with $p_0$ the ray origin, $d_r$ its unit direction and a scalar $t$.

Rays $r$ belonging to a training image $I$ have an associated RGB color $\hat c_r\in\mathbb R^3$, a semantic class $\hat k_r\in\mathcal K$ and a 2D instance ID $\hat h_r \in\mathcal H_I$. The 2D instance ID $\hat h_r$ is defined only for \textit{thing} classes, \ie only if $\hat k_r\in\mathcal K_\mathtt T$, where $\mathcal K_\mathtt T$ and $\mathcal K_\mathtt S$ partition $\mathcal K$ into \textit{thing} and \textit{stuff} classes, respectively.

These semantic and instance labels, generated using a pre-trained 2D segmentation network~\cite{cheng2022masked}, are often noisy and inconsistent, as shown in Fig.~\ref{fig:m2f_inconsistencies}. 
While 2D networks provide a probabilistic distribution over the predicted classes and a confidence per pixel (ray), these are often very peaked, even for incorrectly predicted segments (Fig.~\ref{fig:tta_confidences}). To better estimate  class probabilities and confidences, we run test-time augmentations on the input images (e.g., horizontal flip, scale, brightness, contrast, etc.) and fuse the resulting segmentations by segment clustering (see supplementary for details). This gives us for each  pixel (ray $r$) in the training images, a semantic class distribution $\hat\kappa_r$ over the classes $\mathcal K$ and a confidence estimate $w_r$ of the prediction.

\subsection{Scene Representation and Rendering}
\label{sec:scene_repr}

\mypara{Panoptic radiance field.}
A \emph{panoptic radiance field} is an implicit, volumetric scene representation modeled as a function $\Phi(x,d)$, which assigns to each 3D point $x\in\mathbb R^3$ and viewing direction $d\in \mathbb S^2$ a density $\sigma\in\mathbb R_{\geq 0}$, a semantic class distribution $\kappa$ over $\mathcal K$, a distribution $\pi$ over surrogate identifiers $\mathcal J$ and an RGB color $c\in\mathbb R^3$.  From this representation, we can derive a per-point distribution over 3D object IDs by considering probability $\kappa(k)\pi(j)$ for each 3D object ID $(k,j)\in\mathcal H_\mathtt{3D}$, with $\mathcal H_\mathtt{3D}:=\mathcal K_\mathtt T\times\mathcal J$ being the set of all possible 3D instances our model can produce. The function $\Phi$ leverages a TensoRF~\cite{Chen2022ECCV} representation for color and density, and we introduce two small MLPs to model the semantic and surrogate identifier fields.

\mypara{Volumetric rendering.}
Given the density field $\sigma$ from $\Phi$ as a function of a point $x$, we can render any vector field $f$ over 3D points along a ray $r$ by the rendering equation~\cite{kajiya1984ray}:
\begin{equation}
    R[f|r, \sigma]:=\int_{0}^{\infty}\alpha_t(r)\sigma(r_t)f(r_t, d_r)dt\,,
\end{equation}
where $\alpha_t$ is the transmittance probability at $t$
\begin{equation}
    \alpha_t(r):=\exp\left(-\int_0^t\sigma(r_s)ds\right).
\end{equation}
For brevity, we use $f_r$ to denote the rendered vector field $f$ along ray $r$ when the density field $\sigma$ is clear from the context, \ie $f_r:= R[f|r,\sigma]$.
In particular, this shorthand will be used for rendering all vector-valued fields that are implicitly provided by $\Phi$, namely color $c$, semantic class distributions $\kappa$ and surrogate ID distributions $\pi$. For example, $c_r:= R[c|r,\sigma]$ represents the color field $c$ rendered along ray $r$.

\mypara{Rendering panoptic segmentations.} 
One application of \OURS is rendering 3D-consistent 2D panoptic segmentations from novel views. Given a pixel from a novel view corresponding to ray $r$ in world coordinates, we can compute a rendered semantic class distribution $\kappa_r$ using the rendering equation above and obtain a semantic class $k^\star_r$ for the same pixel as the most probable class, \ie $k^\star_r:=\arg\max_{k\in\mathcal K}\kappa_r(k)$. If $k^\star_r$ is a thing class, we can also compute the most probable surrogate identifier $j^\star_r:=\arg\max_{j\in\mathcal J}\pi_r(j)$ and form a unique 3D instance identifier as the pair $(k^\star_r,j^\star_r)\in\mathcal H_\mathtt{3D}$, which is consistent across the scene. This enables generating multi-view consistent panoptic segmentations for novel views.

\subsection{Loss Functions}
\label{sec:loss_fn}

\mypara{Appearance loss.}
Give a batch of rays $R$, the appearance loss is given as a standard squared Euclidean distance between rendered color field and the ground-truth color:
\begin{equation}
    L_\mathtt{RGB}(R):=\frac{1}{|R|}\sum_{r\in R}\Vert c_r-\hat c_r\Vert^2\,.
\end{equation}

\begin{figure*}
\includegraphics[width=\linewidth]{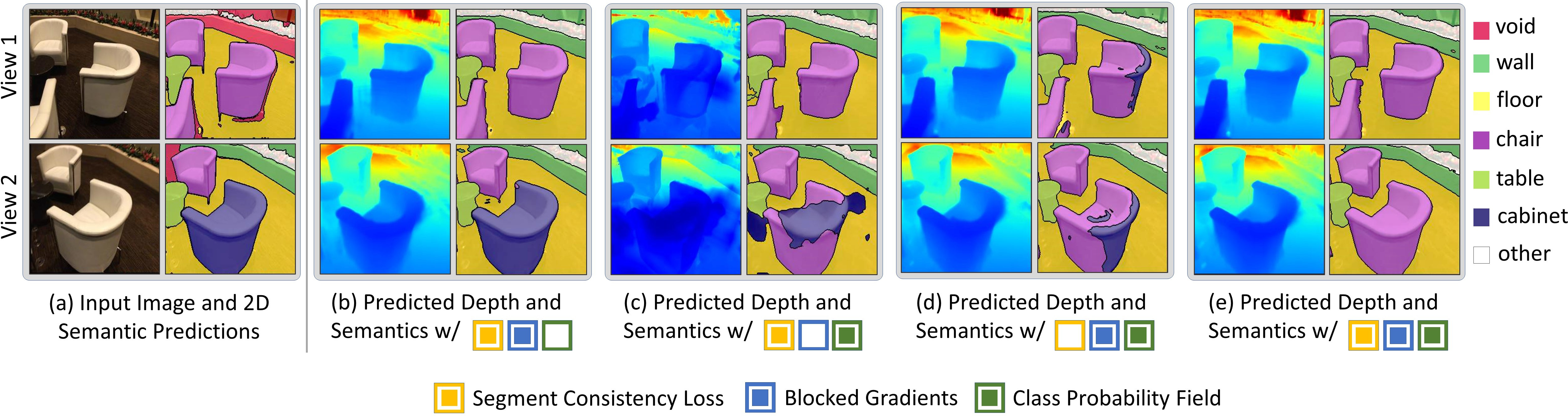}
\vspace{-0.5cm}
\caption{Ablations over components. (a) 2D machine-generated label inconsistency, with the chair labeled as `chair' and `cabinet' in different views. (b) Unbounded segmentation field outputs (e.g. MLP predicts logits) can result in inconsistent predictions even with good depth estimates due to improper distribution along the ray.
%Since volumetric rendering involves a weighted sum of semantic field outputs along the ray, the model can achieve this fitting by making the logits arbitrarily large for some points with non zero weights along the ray for one of the views. 
(c) Even with class probability outputs, the model can change the density (and thereby the rendering weights) to better fit noisy 2D labels. %This can be avoided by not letting the gradients from the segmentation branches affect the density. 
(d) Blocking segmentation gradients along with bounding field output produces a mix of 2 classes for the same chair instance, the optima for the NCE loss on segmentations. Using an additional segment consistency loss improves object consistency. (e) Our final method with all components enabled results in clean and consistent segmentations.}
\label{fig:probability_field_grad_block}
\vspace{-0.30cm}
\end{figure*}

\mypara{Semantic loss.}
As described in Sec.~\ref{sec:input_data}, each ray $r$ has an associated probability distribution $\hat\kappa_r$ over the semantic classes $\mathcal K$ and a prediction confidence $w_r$ obtained as test-time augmented predictions of a 2D pre-trained {panoptic} segmentation model.
Given the semantic distribution field $\kappa_r$ rendered along ray $r$, the semantic loss at $r$ is given by the cross entropy of $\kappa_r$ relative to $\hat \kappa_r$. For a batch of rays $R$,
\begin{equation}
    L_\mathtt{sem}(R):=-\frac{1}{|R|}\sum_{r\in R}w_r\sum_{k\in\mathcal K}\hat\kappa_r(k)\log\kappa_r(k)\,.
\end{equation}
\theoremstyle{remark}
\newtheorem*{remark}{Remark}
\emph{Logits vs. Probabilities.}
Standard implementations of semantic fields render class logits rather then class distributions~\cite{zhi2021place, wang2022dm, nerfstudio, kundu2022panoptic}, converting the rendered logits into probabilities a posteriori via  softmax. This approach can potentially introduce semantic inconsistencies across different viewpoints, since it endows the model with the ability of violating geometric constraints induced by the density. Specifically, due to the unbounded nature of logits, a significant signal could in principle be generated even in low density areas, by just providing sufficiently large logits. We resolve this by instead rendering the bounded probability distribution. See Fig.~\ref{fig:probability_field_grad_block} for an example.

\mypara{Instance loss.}
Let $R$ denote a batch of rays from an image $I$, $R_h$ the subset of rays in $R$ that belong to 2D instance $h\in\mathcal H_I$, and $H_R\subseteq \mathcal H_I$ the subset of 2D instances that are represented in the batch of rays $R$.
Each ray $r$ is assigned the most compatible 3D surrogate identifier via an injective mapping $\Pi^\star_R$, given its 2D machine-generated instance $\hat h_r$. The loss minimizes the log-loss of the corresponding predicted probability averaged over all rays in $R$:
\begin{equation}
    L_\mathtt{ins}(R):=-\frac{1}{|R|}\sum_{r\in R}w_r\log\pi_r(\Pi^\star_R(\hat h_r))\,,
\end{equation}
with $w_r$ as the prediction confidence. The optimal injective mapping $\Pi^\star_R$ is given by
\begin{equation}
    \Pi^\star_R:=\argmax_{\Pi_R}\sum_{h\in H_R}\sum_{r\in R_h}\frac{\pi_r(\Pi_I(h))}{|R_h|}\,,
\end{equation}
which can be solved as a linear assignment problem.

\mypara{Segment consistency loss.}
Let $R$ denote a batch of rays from an image, partitioned into groups of rays $\{R_1,...,R_m\}$ based on the panoptic segment each ray belongs to. That is, rays are clustered based on their 2D instance or \textit{stuff} class. The segment consistency loss for $R$ is:
\begin{equation}
    L_\mathtt{con}(R):=-\frac{1}{|R|}\sum_{r\in R}w_r\sum_{i=1}^m  \log \kappa_r(K_i)\,,
\end{equation}
where $w_r$ is the prediction confidence and $K_i$ is the most probable predicted semantic class within the group $R_i$, \ie
$
    K_i:= \argmax_{k\in\mathcal K}\sum_{r\in R_i}w_r\kappa_r(k)\,.
$

\mypara{Training objective.}
Let $R=R_{S}\cup R_{I}$ be a batch of rays with $R_S$ as rays randomly sampled across the scene and $R_I$ as rays sampled from a single image, then the total loss over the network parameters is 
$
% \label{eqn:total_loss}
    % \begin{aligned}
    L_\mathtt{tot}(R) := \lambda_{\mathtt{ins}} L_\mathtt{ins} (R_I) + \lambda_{\mathtt{con}} L_\mathtt{con}(R_I) + \lambda_{\mathtt{RGB}} L_\mathtt{RGB}(R_S)  + \lambda_{\mathtt{sem}} L_\mathtt{sem}(R_S).
    % \end{aligned}
$

\vspace{1mm}
Note that it is important that the supervision for the segmentation with noisy 2D machine labels not influence the geometry. Otherwise the model can still satisfy inconsistent labels by changing the rendering weights, resulting in cloudy geometry to satisfy conflicting segmentation labels (Fig.~\ref{fig:probability_field_grad_block}c). To avoid this issue, we stop gradients from the segmentation branches back to the densities.

\mypara{Implementation Details.} Our model is implemented with Pytorch and trained using Adam~\cite{kingma2014adam} with learning rates of $5\times10^{-4}$ for the MLPs and $1\times10^{-2}$ for the TensoRF lines and planes. For $L_{\mathtt{tot}}$ we use $\lambda_{\mathtt{con}}=1.35$, $\lambda_{\mathtt{ins}}=\lambda_{\mathtt{sem}}=\lambda_{\mathtt{RGB}}=1$. We train on 1 NVIDIA A6000 for 450k iterations ($\sim 10$ hours). Further details are in the supplementary.

%------------------------------------------------------------------------

\section{Experiments}
\label{sec:experiments}
We evaluate our method on the tasks of novel view synthesis and novel view panoptic segmentation, and further show scene editing as a possible application.

{
\begin{table*}[tp]
\begin{center}
\small
\resizebox{\textwidth}{!}{
\begin{tabular}{l|lll|lll|lll}  
\toprule
\multirow{2}{*}{Method} & \multicolumn{3}{c|}{HyperSim~\cite{roberts2021hypersim}} & \multicolumn{3}{c|}{Replica~\cite{straub2019replica}} & \multicolumn{3}{c}{ScanNet~\cite{dai2017scannet}}  \\ 
& mIoU$\uparrow$ & PQ$^\text{scene}\uparrow$ & PSNR$\uparrow$ &  mIoU$\uparrow$ & PQ$^\text{scene}\uparrow$ & PSNR$\uparrow$  &  mIoU$\uparrow$ & PQ$^\text{scene}\uparrow$ & PSNR$\uparrow$  \\
\midrule
Mask2Former~\cite{cheng2022masked} & 
53.9$_{\;\text{\textcolor{gray}{(-13.9)}}}$& --%40.4$_{\;\text{\textcolor{gray}{(-19.7)}}}$
& -- & 52.4$_{\;\text{\textcolor{gray}{(-14.8)}}}$ & --
%40.8$_{\;\text{\textcolor{gray}{(-17.1)}}}$
& -- & 46.7$_{\;\text{\textcolor{gray}{(-18.5)}}}$ & --%30.4$_{\;\text{\textcolor{gray}{(-28.5)}}}$
& -- \\
SemanticNeRF~\cite{zhi2021place} & 58.9$_{\;\text{\textcolor{gray}{(-8.9)}}}$ & -- & 26.6 & 58.5$_{\;\text{\textcolor{gray}{(-8.7)}}}$ & -- & 24.8 & 59.2$_{\;\text{\textcolor{gray}{(-6)}}}$ & -- & 26.6\\
DM-NeRF~\cite{wang2022dm} & 57.6$_{\;\text{\textcolor{gray}{(-10.2)}}}$ & 51.6$_{\;\text{\textcolor{gray}{(-8.5)}}}$ & 28.1 & 56.0$_{\;\text{\textcolor{gray}{(-11.2)}}}$ & 44.1$_{\;\text{\textcolor{gray}{(-13.8)}}}$ & 26.9 & 49.5$_{\;\text{\textcolor{gray}{(-15.7)}}}$ & 41.7$_{\;\text{\textcolor{gray}{(-17.2)}}}$ & 27.5\\
PNF~\cite{kundu2022panoptic} & 50.3$_{\;\text{\textcolor{gray}{(-17.5)}}}$ & 44.8$_{\;\text{\textcolor{gray}{(-15.3)}}}$ & 27.4 & 51.5$_{\;\text{\textcolor{gray}{(-15.7)}}}$ & 41.1$_{\;\text{\textcolor{gray}{(-16.8)}}}$ & 29.8 & 53.9$_{\;\text{\textcolor{gray}{(-11.3)}}}$ & 48.3$_{\;\text{\textcolor{gray}{(-10.6)}}}$ & 26.7\\
PNF + GT Bounding Boxes & 58.7$_{\;\text{\textcolor{gray}{(-9.1)}}}$ & 47.6$_{\;\text{\textcolor{gray}{(-12.5)}}}$ & 28.1 & 54.8$_{\;\text{\textcolor{gray}{(-12.4)}}}$ & 52.5$_{\;\text{\textcolor{gray}{(-5.4)}}}$ & \textbf{31.6} & 58.7$_{\;\text{\textcolor{gray}{(-6.5)}}}$ & 54.3$_{\;\text{\textcolor{gray}{(-4.6)}}}$ & 26.8\\
\midrule
\OURS & \textbf{67.8} & \textbf{60.1} & \textbf{30.1}& \textbf{67.2} & \textbf{57.9} & 29.6 & \textbf{65.2} & \textbf{58.9} & \textbf{28.5}\\
\bottomrule
\end{tabular}}
    \vspace{-0.1cm}
\caption{Quantitative comparison on novel views from the test set. We outperform both 2D and 3D NeRF methods on semantic and panoptic segmentation tasks. Note that, compared to other methods, \textit{PNF+GT Bounding Boxes} is given the advantage of using ground-truth 3D detections.
%Mask2Former does not preserve object identities across frames, which is especially penalized by our PQ$^\text{scene}$ metric.
Mask2Former does not predict scene-level object instances, thus it can't be evaluated for PQ$^\text{scene}$.}
\label{tab:quantative_evaluation}
\end{center}
\vspace{-0.45cm}
\end{table*}
}
\begin{table}[t]
    \begin{center}
    \resizebox{\linewidth}{!}{
    \begin{tabular}{cccc|ccc}
    \toprule
     \makecell{Segment\\Consistency}  & \makecell{TTA} & \makecell{Bounded\\ Segm. Field} & \makecell{Gradient\\Blocking} & mIoU$\uparrow$ & PQ$^\text{scene}\uparrow$ & PSNR$\uparrow$\\
    \midrule
    \cellcolor{gray!25}\ding{55} & \cellcolor{gray!25}\ding{55} & \cellcolor{gray!25}\ding{55} & \cellcolor{gray!25}\ding{55} & \cellcolor{gray!25}57.3 & \cellcolor{gray!25}47.9 & \cellcolor{gray!25}27.1 \\
    
    \ding{55} & \ding{51} & \ding{51} & \ding{51}  & 60.9  &  54.3 & 28.3\\
    
    \cellcolor{gray!25}\ding{51} & \cellcolor{gray!25}\ding{55} & \cellcolor{gray!25}\ding{51} & \cellcolor{gray!25}\ding{51} & \cellcolor{gray!25}63.1 & \cellcolor{gray!25}55.2 & \cellcolor{gray!25}28.4 \\
    
    \ding{51} & \ding{51} & \ding{55} & \ding{51} & 62.9 & 52.5 & 28.4\\
    
    \cellcolor{gray!25}\ding{51} & \cellcolor{gray!25}\ding{51} & \cellcolor{gray!25}\ding{51} & \cellcolor{gray!25}\ding{55} & \cellcolor{gray!25}61.6 & \cellcolor{gray!25}53.7 & \cellcolor{gray!25}27.2 \\
    
    \ding{51} & \ding{51} & \ding{51} & \ding{51} & \textbf{65.2} & \textbf{58.9} & \textbf{28.5} \\
    
    \bottomrule
    \end{tabular}
    }
    \vspace{-0.2cm}
    \caption{Ablations of our design choices on ScanNet~\cite{dai2017scannet}. The segment consistency loss, test-time augmentations (TTA), probability field, and blocked segmentation gradients  all contribute to robustness against real-world noisy labels.}
    \label{tab:ablations}
    \end{center}
    \vspace{-0.8cm}
\end{table}

\mypara{Data.} We show results on scenes from three public datasets: Hypersim~\cite{roberts2021hypersim}, Replica~\cite{straub2019replica} and ScanNet~\cite{dai2017scannet}. For each of the datasets, we use the available ground-truth poses. The ground-truth semantic and instance labels, however, are only used for evaluation, and are not used for training or refinement of any models. We also show results on in-the-wild scenes, captured with a Pixel3a Android smartphone with COLMAP~\cite{schonberger2016structure}-computed poses. To obtain the machine-generated 2D panoptic segmentations, we use the same publicly available pre-trained model~\cite{cheng2022masked} for all scenes across all datasets. Since this model was trained originally on COCO~\cite{lin2014microsoft}, we map COCO panoptic classes to 21 ScanNet classes (9 \textit{thing} + 12 \textit{stuff}). Hypersim and Replica labels are also mapped to the same class set for evaluation. For the in-the-wild captures we use 31 ScanNet classes (17 \textit{thing} + 14 \textit{stuff}). Further details about the scenes, classes and data splits can be found in the supplementary.

\mypara{Metrics.}
We measure the visual fidelity of the synthesized novel views, and the accuracy of their predicted semantic labels using peak signal to noise ratio (PSNR), and mean intersection over union (mIoU), respectively.
Recent NeRF-based semantic scene modeling works~\cite{fu2022panoptic,kundu2022panoptic} evaluate their panoptic predictions using the Panoptic Quality (PQ)~\cite{kirillov2019panoptic} metric.
Standard PQ, however, does not measure whether instance identities are consistently preserved across views.
To overcome this limitation, we propose a scene-level PQ metric, denoted as PQ$^\text{scene}$.

\mypara{Scene-level Panoptic Quality.}
Given a set of predicted segments $p\in P$, and ground truth segments $g\in G$, the PQ metric is defined by comparing all possible pairs of segments belonging to an image, marking them as a match if $\text{IoU}(p,g)>0.5$ (see Sec.~4.1-4.2 of~\cite{kirillov2019panoptic}).
In PQ$^\text{scene}$, we modify the matching process to work on a \emph{scene level}.
To achieve this, we first merge the segments into subsets $\mathcal{P}\subset P$, $\mathcal{G}\subset G$ which contain all segments that, for a certain scene, belong (or are predicted to belong) to the same instance or stuff class.
Then, we compare all pairs of \emph{subsets} for each scene, and record a match when $\text{IoU}(\mathcal{P},\mathcal{G})>0.5$.%
\footnote{Note that PQ$^\text{scene}$ can be trivially implemented by evaluating standard PQ after tiling together the predictions and ground truths of each scene.}

\begin{figure}
\begin{center}
\includegraphics[width=0.95\linewidth]{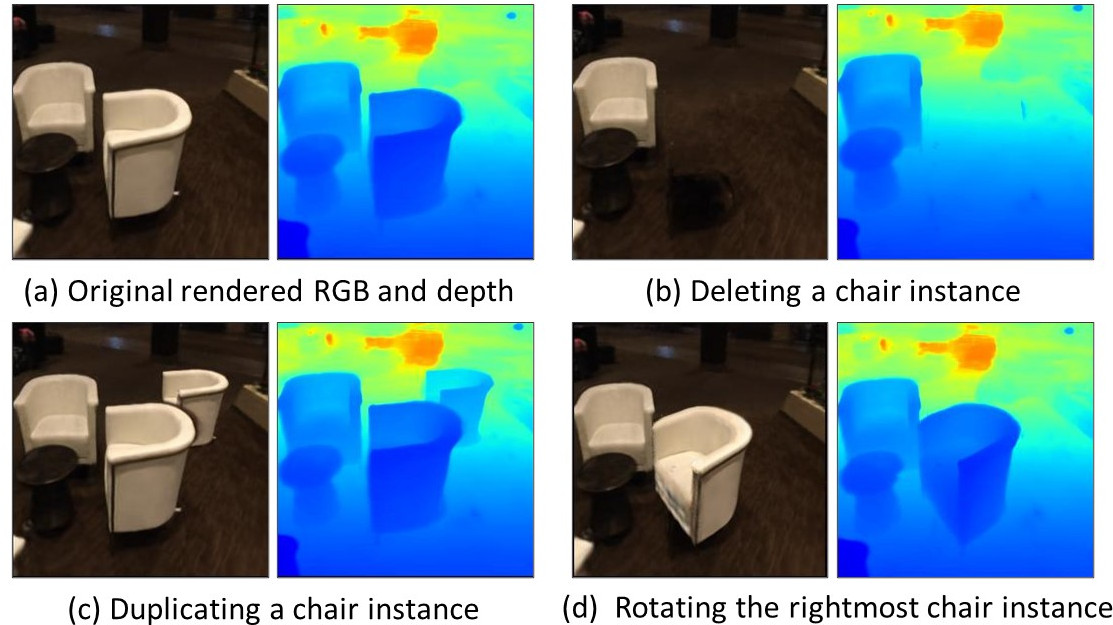}
\vspace{-0.1cm}
\caption{Scene Editing. Once trained, we can generate novel views of a scene with object instances (b) deleted, (c) duplicated or (d) manipulated with affine transforms.
}
\label{fig:experiments_editing}
\end{center}
\vspace{-0.5cm}
\end{figure}

\begin{figure*}
\includegraphics[width=\linewidth]{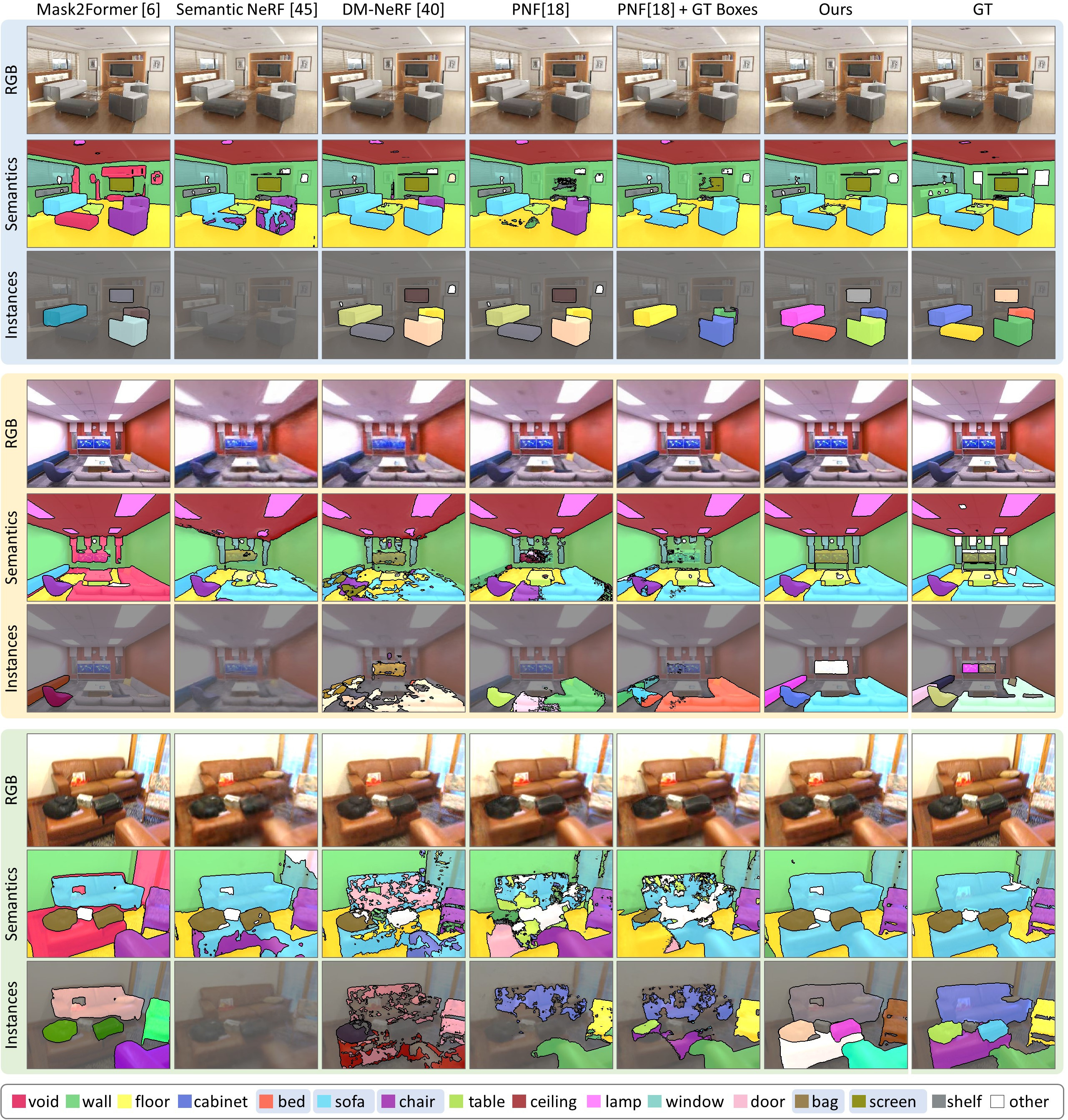}
\vspace{-0.50cm}
\caption{Novel views and their corresponding semantics and instances on \colorbox{blue!10}{Hypersim}~\cite{roberts2021hypersim}, \colorbox{yellow!30}{Replica}~\cite{straub2019replica}, and \colorbox{green!10}{ScanNet}~\cite{dai2017scannet} (top to bottom, respectively). Highlighted legend labels represent instanced (\textit{thing}) classes. All 3D methods use the same posed RGB images and 2D machine-generated labels for training. PNF~\cite{kundu2022panoptic} additionally uses 3D bounding box predictions, while PNF + GT Boxes uses ground truth bounding boxes for instance classes. We outperform the state of the art, producing clean and consistent segmentation masks.}
\vspace{-0.10cm}
\label{fig:experiments_dataset}
\end{figure*}

\begin{figure*}
\vspace{-0.4cm}
\begin{center}
\includegraphics[width=0.90\linewidth]{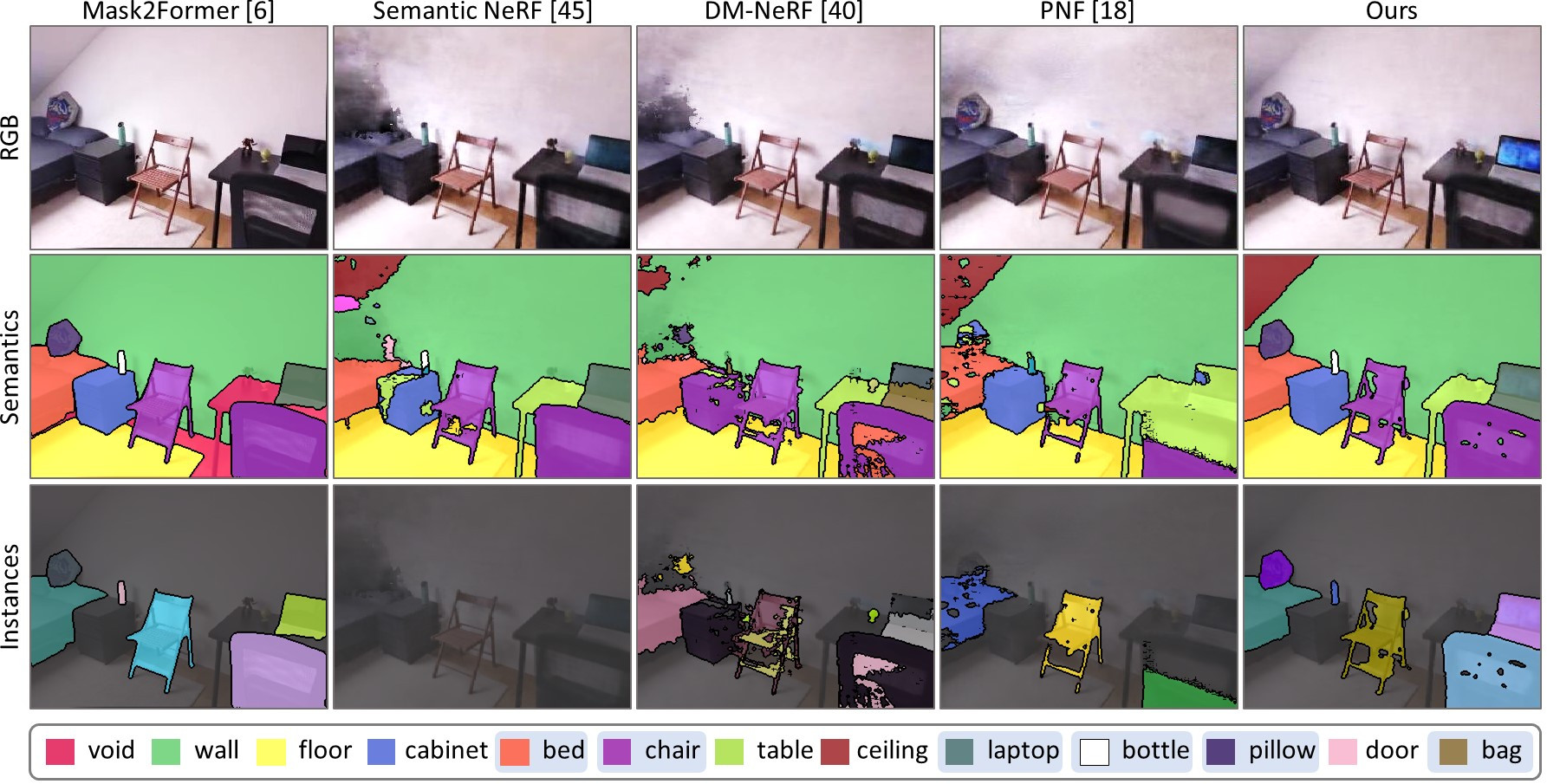}
\vspace{-0.2cm}
\caption{Novel views and their corresponding semantics and instances on \textit{in-the-wild} room capture. Highlighted labels in the legend represent instanced (\textit{thing}) classes. All 3D methods use the same posed RGB images and 2D machine-generated labels for training. PNF~\cite{kundu2022panoptic} additionally uses predictions from a 3D bounding box detector.}
\label{fig:experiments_itw}
\end{center}
\vspace{-0.4cm}
\end{figure*}

\subsection{Results}

We compare with state-of-the-art 2D and NeRF-based 3D semantic and panoptic segmentation methods:
Mask2Former~\cite{cheng2022masked} predicts 2D panoptic segmentation, SemanticNeRF~\cite{zhi2021place} predicts 3D semantic segmentation, and Panoptic Neural Fields (PNF)~\cite{kundu2022panoptic} and DM-NeRF~\cite{wang2022dm} predict 3D panoptic segmentation.
Our method and the neural field baselines are all trained using the same set of images, poses and Mask2Former generated 2D labels.
For PNF, we additionally provide bounding boxes from a state-of-the-art multi-view 3D object detector~\cite{rukhovich2022imvoxelnet} pre-trained on ScanNet, or taken from the ground truth (PNF + GT Bounding Boxes).
More details are provided in the supplementary.

As shown in Tab.~\ref{tab:quantative_evaluation} and Fig.~\ref{fig:experiments_dataset}, we outperform baselines across all datasets on both semantic and panoptic segmentation tasks, without sacrificing view synthesis quality.
Fig.~\ref{fig:experiments_itw} additionally shows qualitative results on an in-the-wild capture.
We further demonstrate improved consistency and segmentation quality over baselines in the supplemental video.

Our method significantly improves over Mask2Former, harmonizing its inconsistent and noisy outputs by lifting them to 3D.
This is reflected by the IoU scores in Tab.~\ref{tab:quantative_evaluation} (-13.9\% \wrt \OURS), and by evaluating standard PQ on ScanNet, where we register a scores of 43.6\% and 60.4\% for Mask2Former and our method, respectively.
While SemanticNeRF was shown to be robust to synthetic pixel noise, we find that it struggles to handle the error patterns of machine-generated labels.
Similarly, DM-NeRF seems to suffer with segment fragmentation when confronted with real-world scenes and machine-generated panoptic labels (Fig.~\ref{fig:experiments_dataset}).
Finally, PNF's sensitivity to errors from the 3D bounding box detector is well highlighted in Tab.~\ref{tab:quantative_evaluation} (-15.7 to -10.6\% PQ$^\text{scene}$ \wrt \OURS).
When provided with g.t. boxes, PNF is able to partially close the PQ$^\text{scene}$ gap, but \OURS still shows greater robustness to noise in 2D machine-generated labels.

\subsubsection{Ablations} 

In Tab.~\ref{tab:ablations}, we show a set of ablations on the ScanNet data.
As a first baseline (row 1), we disable all of the robustness-oriented design choices described in Sec.~\ref{sec:loss_fn}, obtaining a substantial drop in performance (-8\% mIoU, -11\% PQ$^\text{scene}$).

\mypara{Segment consistency loss.} 
Fig.~\ref{fig:probability_field_grad_block}(d) shows how inconsistent 2D segmentations under different views result in a blend of two labels for the same physical object.
This effect is counteracted by our segment consistency loss: disabling it causes the largest drop in mIoU (Tab.~\ref{tab:ablations}, row 2).

\mypara{Test-time augmentation.}
As shown in Fig.~\ref{fig:tta_confidences}, Mask2Former predictions tend to be highly confident, even for incorrectly predicted classes.
Test-time augmentation provides smoother confidence estimates and improved masks, boosting both semantic metrics (Tab.~\ref{tab:ablations}, row 3).

\mypara{Bounded vs. unbounded segmentation field.}
Predicting unbounded logits from the segmentation branch, as in other recent NeRF-based semantic scene modeling works~\cite{zhi2021place,wang2022dm,kundu2022panoptic,nerfstudio}, allows the model to ``cheat'' and predict inconsistent labels even with accurate geometry (see \cref{fig:probability_field_grad_block}(a,b), \cref{sec:loss_fn}).
Switching to bounded segmentation fields drastically improves consistency, as shown by the PQ$^\text{scene}$ score in Tab.~\ref{tab:ablations}, row 4.

\mypara{Blocking semantics-to-geometry gradients.}
Even with a bounded segmentation field, the model can be pushed to learn wrong geometry in an attempt to satisfy inconsistent 2D labels, as shown in Fig.~\ref{fig:probability_field_grad_block}(c).
This is reflected in a degradation in view synthesis PSNR (Tab.~\ref{tab:ablations}, row 5), which we can solve by blocking gradients from semantic and instance branches back to geometry.
 
\subsubsection{Scene Editing} In addition to generating novel views and their panoptic masks, \OURS can be used for scene editing. 
As shown in Fig.~\ref{fig:experiments_editing}, once trained, we can generate novel views of a scene with object instances deleted, duplicated or manipulated under affine transformations. Given an instance label, deletion is achieved by setting the density of points where a predicted instance is equal to the instance to be deleted to be zero. For duplication, given a new position and rotation for an object to be duplicated, we manipulate the rays passing through the resulting region to query the original instance region instead. Manipulation is achieved by combining deletion and duplication.

\subsubsection{Limitations}
\OURS shows considerable improvements over the state of the art; however, several limitations remain.
Our method uses predictions from a pre-trained panoptic segmentation model, and hence is limited to classes with which the original model was trained.
In this context, it would be interesting to explore open world segmentation~\cite{li2022language,ghiasi2021open} with self-supervised instance clustering.
Similar to other NeRF-based approaches, our method is currently run off-line due to lengthy pre-processing for pose estimation, 2D segmentation inference, and neural field optimization; here, a promising avenue would be to integrate our approach with state-of-the-art SLAM approaches that run in real-time.

%-------------------------------------------------------------------------
\section{Conclusion}
\label{sec:conclusions}
We have introduced \OURS, a novel approach to lift 2D machine-generated panoptic labels to an implicit 3D volumetric representation. As a result, our model can produce clean, coherent, and 3D-consistent panoptic segmentation masks together with color and depth images for novel views. Compared to state of the art, our model is more robust to the inherent noise present in machine-generated labels, hence resulting in significant improvements across datasets while providing the ability to work on in-the-wild scenes. We believe \OURS is an important step towards enabling more robust, holistic 3D scene understanding while posing minimal input requirements.

\section*{Acknowledgements}
This work was done during Yawar’s and Norman's internships at Meta Reality Labs Zurich as well as at TUM, funded by a Meta SRA.
Matthias Nie{\ss}ner was also supported by the ERC Starting Grant Scan2CAD (804724). Angela Dai was supported by the Bavarian State Ministry of Science and the Arts coordinated by the Bavarian Research Institute for Digital Transformation (BIDT).
We would like to thank Justus Thies, Artem Sevastopolsky, and Guy Gafni for helpful discussions and their feedback.

% %%%%%%%%% REFERENCES
{\small
\bibliographystyle{ieee_fullname}
\bibliography{egbib}
}
\newpage
\begin{appendix}
In this supplementary document, we discuss additional details about our method \OURS.
Specifically. in Section~\ref{sec:supp_tta} we give additional details about our test time augmentation algorithm. 
A comparison of rendering performance of our method compared to the baselines is reported in Section~\ref{sec:supp_renderperf}.
We also provide implementation details of our method and the baselines (Section~\ref{sec:supp_impldeets}), and the data used for experiments in the main paper (Section~\ref{sec:supp_data}).
Finally, we report additional metrics, scene Segmentation Quality (SQ$^{\mathtt{scene}}$) and Retrieval Quality (RQ$^{\mathtt{scene}}$), in Section~\ref{sec:supp_results}.

\section{Test-time Augmentation for Mask2Former}
\label{sec:supp_tta}

In this section we describe the test-time augmentation strategy we adopt to obtain improved panoptic segmentation masks and per-pixel confidence scores from Mask2Former~\cite{cheng2022masked}.

\subsection{Test-time Augmentation}
We run a pre-trained Mask2Former network on multiple augmented versions of each input image, using the following set of transformations: horizontal flip, rescale, contrast, RGB-shift, random gamma, random brightness \& contrast, median blur, sharpen, and arbitrary combination of the previously mentioned augmentations.
For each transformation, we intercept the Mask2Former outputs before its ``panoptic fusion'' stage, \ie right after the transformer and pixel decoders (see Sec.3 of~\cite{cheng2022masked} for details).
These outputs consist of a set of candidate segments, represented as 2D soft masks paired with probability distributions over the classes.
After transforming the candidate segments back to the original image resolution and orientation, our next objective is to fuse them into a single, coherent panoptic segmentation.

\subsection{Fusing Mask2Former predictions}
We denote the candidate segments predicted from all augmented versions of the image as a set of pairs $(\mathsf{m}_i, \mathbf{p}_i), i=1,\dots,N$, where $\mathsf{m}_i(x, y)\in[0, 1]$ is the predicted probability of pixel $(x, y)$ to belong to segment $i$, and $\mathbf{p}_i =[p_i^1, ..., p_i^C]$ is the segment's predicted probability distribution over $C$ classes.
In the following, we describe a mechanism to combine these predictions into a single panoptic segmentation with associated confidences, following three steps: segment clustering, cluster aggregation and panoptic fusion.

\mypara{Segment clustering.}
We build a graph $(\mathcal{V}, \mathcal{E})$, where $\mathcal{V}=\{1,...,N\}$, and $\mathcal{E}=\{(i, j)| i,j \in \mathcal{V} \wedge s(i, j) \geq \theta \}$.
The matching function $s(i, j)$ is defined as a ``soft-IoU''
\begin{equation*}
s(i, j) = \frac{
    \sum_{x,y} \min (\mathsf{m}_i(x, y), \mathsf{m}_j(x, y))
}{
    \sum_{x,y} \max (\mathsf{m}_i(x, y), \mathsf{m}_j(x, y))
}\,,
\end{equation*}
and $\theta$ is a matching threshold (e.g. $\theta=0.5$).
In other words, we add an edge between two segments if their soft-IoU is greater than $\theta$.
By finding the connected component of this graph, we partition the segments into clusters $\mathcal{K}\subset\mathcal{V}$.

\mypara{Cluster aggregation.}
After clustering the segments, we define a new set of masks and class probabilities, this time associated with clusters instead of segments.
We denote these as $\mathsf{\hat{m}}_\mathcal{K}(x, y)$ and $\mathbf{\hat{p}}_\mathcal{K}=[\hat{p}_\mathcal{K}^1, ..., \hat{p}_\mathcal{K}^C]$, respectively, and compute them by simply averaging the masks and probabilities of all segments belonging to each cluster

\begin{align*}
\mathsf{\hat{m}}_\mathcal{K}(x, y) &= \frac{1}{|\mathcal{K}|}
    \sum_{i\in \mathcal{K}} \mathsf{m}_i(x, y)\,, \\
\mathbf{\hat{p}}_\mathcal{K} &= \frac{1}{|\mathcal{K}|}
    \sum_{i\in \mathcal{K}} \mathbf{p}_i\,.  
\end{align*}

\mypara{Panoptic fusion.}
Given this new set of masks and class probabilities, we fuse them into a single overall panoptic prediction with an algorithm akin to the one used in the final stage of~\cite{cheng2022masked}.
Specifically, we follow these steps:
\begin{enumerate}
    \item For each cluster $\mathcal{K}$, we determine the most likely class $c_\mathcal{K}^*=\arg\max_c \hat{p}_\mathcal{K}^c$, and the corresponding probability $p_\mathcal{K}^*=\max_c \hat{p}_\mathcal{K}^c$.
    \item We scale $\mathcal{K}$'s mask by $p_\mathcal{K}^*$ to obtain $\mathsf{\bar{m}}_\mathcal{K}(x, y) = p_\mathcal{K}^* \mathsf{\hat{m}}_\mathcal{K}(x, y)$.
    \item We assign image pixels to clusters with the rule: $(x, y)$ is assigned to $\mathsf{k}^*(x, y) = \arg\max_\mathcal{K} \mathsf{\bar{m}}_\mathcal{K}(x, y)$, and its confidence is set to $\mathsf{s}(x,y) = \max_\mathcal{K} \mathsf{\bar{m}}_\mathcal{K}(x, y)$.
\end{enumerate}
At the end of this process, each pixel will have a class $c_{\mathsf{k}^*(x, y)}^*$ and a confidence $\mathsf{s}(x,y)$.
Furthermore, pixels of thing classes can be partitioned into instances according to their cluster assignment $\mathsf{k}^*(x, y)$.

\section{Rendering Performance}
\label{sec:supp_renderperf}
Tab.~\ref{tab:supp_performance} compares the time taken to render a batch of 2048 rays for each method on an NVIDIA RTX A6000 GPU. Due to the hybrid representation from TensoRF, our model delivers a faster rendering performance compared to the baselines. 

{
\begin{table}
\begin{center}
\small
{
\begin{tabular}{l|r}  
\toprule
Method & Time to render 2048 rays  \\ 
\midrule
PNF~\cite{kundu2022panoptic} & 119.7 ms\\
DM-NeRF~\cite{wang2022dm} & 66.5 ms\\
Semantic-NeRF~\cite{zhi2021place} & 65.7 ms \\
\OURS (Ours) & \textbf{13.1} ms\\
\bottomrule
\end{tabular}}
\caption{Time taken to render a batch of 2048 rays on a NVIDIA RTX A6000 GPU.}
\label{tab:supp_performance}
\end{center}
\end{table}
}
{
\begin{table}
\begin{center}
\small
\resizebox{\linewidth}{!}
{
\begin{tabular}{l|l|l|l}  
\toprule
Method & {HyperSim~\cite{roberts2021hypersim}} & {Replica~\cite{straub2019replica}} & {ScanNet~\cite{dai2017scannet}}  \\ 
\midrule
Mask2Former~\cite{cheng2022masked} & 50.52 & 50.10 & 43.6\\
\OURS (Ours) & \textbf{66.84} & \textbf{63.79} & \textbf{60.4}\\
\bottomrule
\end{tabular}
}
\caption{Conventional PQ scores on novel views from the test set.}
\label{tab:supp_pq}
\end{center}
\end{table}
}

{
\begin{table*}[tp]
\begin{center}
\small
\resizebox{\textwidth}{!}
{
\begin{tabular}{l|ll|ll|ll}  
\toprule
\multirow{2}{*}{Method} &
\multicolumn{2}{c|}{HyperSim~\cite{roberts2021hypersim}} &
\multicolumn{2}{c|}{Replica~\cite{straub2019replica}} &
\multicolumn{2}{c}{ScanNet~\cite{dai2017scannet}}  \\  

& SQ$^\text{scene}\uparrow$ & RQ$^\text{scene}\uparrow$ & SQ$^\text{scene}\uparrow$ & RQ$^\text{scene}\uparrow$  & SQ$^\text{scene}\uparrow$ & RQ$^\text{scene}\uparrow$ \\

\midrule

DM-NeRF~\cite{wang2022dm} & 62.06 & 55.45 & 58.68 & 47.68 & 53.26 & 46.13 \\
PNF~\cite{kundu2022panoptic}  & 55.33 &	47.51 & 53.62  &	44.10  & 62.96 & 50.73 \\
PNF~\cite{kundu2022panoptic} + GT Bounding Boxes& 68.23	 & 53.35  & 62.15  & 50.81  & 70.01	& 55.87\\

\midrule

\OURS (Ours)  & \textbf{70.35} & \textbf{64.32} & \textbf{69.10} & \textbf{63.61}& \textbf{73.50} & \textbf{64.95} \\ 
\bottomrule
\end{tabular}
}
\caption{SQ and RQ metrics for on novel views from the test set.}
\label{tab:supp_results}
\end{center}
\end{table*}
}

\section{Implementation Details}
\label{sec:supp_impldeets}
\subsection{Panoptic Lifting}
\OURS uses TensoRF~\cite{Chen2022ECCV} for modeling the scene density and radiance. Specifically, we use the Vector-Matrix (VM) decomposition, with number of density and appearance components set to 16 and 48 respectively. The starting grid resolution is set to $128^3$ and goes upto $192^3$ at the end of the optimization. 27 color features are decoded with a tiny 2 layer MLP with positional encoding with 2 components to encode the view direction and the features. 

To model the semantic class distribution and surrogate identifiers we make use of two small view-independent MLPs. The semantic MLP has 5 layers with a width of 256 and outputs a probability distribution over the target classes for any given input position. The surrogate identifier is a 3 layer MLP which generates a distribution over max $k$ identifiers (set to 50 in our experiments). 
%These predicted surrogate identifiers in combination with the predicted classes determine the instance labels for the \textit{thing} classes. 
Neither of these MLPs use positional encoding. We choose to go with MLPs instead of Vector-Matrix decompositions for semantics and surrogate identifiers for memory size constraints. Our model is trained with a batch of 2048 rays, with a learning rate of 0.0005 for MLPs and 0.02 for the TensoRF lines and planes.

\subsection{Baselines}
We use the publicly available Mask2Former~\cite{cheng2022masked} code and models, without any retraining or fine-tuning. For all methods that use Mask2Former instance labels (including ours), instance counts are renumbered to be distinct across frames. For Semantic-NeRF~\cite{zhi2021place} and DM-NeRF~\cite{wang2022dm}, we use their publicly released code. Since DM-NeRF outputs the labels as abstract instance identifiers, we create a map from instance to class using the instance's majority class across the train set as its assigned class. 

Since Panoptic Neural Fields~\cite{kundu2022panoptic} does not provide a public implementation, we re-implement it based on details from the paper. We do not use their prior-based initialization since it requires additional 3D datasets for the instanced classes.  In the original implementation, PNF uses a monocular 3D detector, which is essential when dealing with dynamic objects varying across frames. However, since the task here deals with static scene, it is more fair to use a multi-view detector for getting the bounding boxes. We use a state-of-the-art multiview detector~\cite{rukhovich2022imvoxelnet} pretrained on ScanNet for getting object bounding boxes for PNF in our experiments. Note that for getting a reasonable 3D detector performance, it is required that the camera poses are scaled and centered similarly to the original ScanNet training data. We perform these pose corrections for Replica~\cite{straub2019replica}, Hypersim~\cite{roberts2021hypersim} and in-the-wild scenes. Since this correction requires an estimate of scale, we use for pose correction the ground-truth depth from Replica and Hypersim, and NeRF optimized depth for scenes in the wild. We further show result with a variant of PNF that uses ground-truth detections, except for in-the-wild data where not ground-truth is available.

All models (including ours) are trained with Mask2Former~\cite{cheng2022masked} generated labels. 

\section{Data}
\label{sec:supp_data}

Tab.~\ref{tab:supp_scene_info} shows the scenes and their corresponding number of frames. The available posed images are split into 75\% views for training and 25\% intermediately sampled test views. Note that for each of the datasets, the ground-truth semantic and instance labels are only used for evaluation, and are not used for training or refinement of any models. 

Since the original model ({swin\_large\_IN21k}) was trained on COCO~\cite{lin2014microsoft}, and the labels for evaluation come from different datasets, we map the Mask2Former predictions as well as the ground-truth labels across all the datasets used in our experiments to ScanNet 21 classes (Tab.~\ref{tab:supp_classes} left). For in the wild scenes, we use 31 ScanNet classes listed in Tab.~\ref{tab:supp_classes} (right).

\begin{table}[t]
    \begin{center}
    {
    \begin{tabular}{l|l|r}
    \toprule
        Dataset & Scene & \# Frames\\
    \midrule
        HyperSim & ai\_001\_003 & 100 \\ 
        HyperSim & ai\_001\_008 & 100 \\ 
        HyperSim & ai\_001\_010 & 300 \\ 
        HyperSim & ai\_008\_004 & 63 \\ 
        HyperSim & ai\_010\_005 & 100 \\ 
        HyperSim & ai\_035\_001 & 200 \\ 
        ScanNet & scene0050\_02 & 874 \\ 
        ScanNet & scene0144\_01 & 678 \\ 
        ScanNet & scene0221\_01 & 780 \\ 
        ScanNet & scene0300\_01 & 929 \\ 
        ScanNet & scene0354\_00 & 563 \\ 
        ScanNet & scene0389\_00 & 708 \\ 
        ScanNet & scene0423\_02 & 855 \\ 
        ScanNet & scene0427\_00 & 659 \\ 
        ScanNet & scene0494\_00 & 740 \\ 
        ScanNet & scene0616\_00 & 758 \\ 
        ScanNet & scene0645\_02 & 726 \\ 
        ScanNet & scene0693\_00 & 866 \\ 
        Replica & office\_0 & 900 \\ 
        Replica & office\_2 & 900 \\ 
        Replica & office\_3 & 900 \\ 
        Replica & office\_4 & 900 \\ 
        Replica & raw & 900 \\ 
        Replica & room\_0 & 900 \\ 
        Replica & room\_1 & 900 \\ 
        Replica & room\_2 & 900 \\ 
        In the wild & office & 1100 \\ 
        In the wild & bed\_room & 1100 \\ 
        In the wild & meeting\_room & 1100 \\ 
    \bottomrule
    \end{tabular}
    }
    \caption{Scenes used for evaluations in our experiments. Note that the in the wild scenes are only used for qualitative evaluation (shown in the supplementary video) since ground truth labels are not available for a qualitative comparison with baselines.}
    \label{tab:supp_scene_info}
    \end{center}
    \vspace{-0.8cm}
\end{table}

\begin{table}[t]
    \begin{center}
    {
    \begin{tabular}{l|l}
    \toprule
        Class & Type \\
    \midrule
        wall &	Stuff \\
        floor &	Stuff \\
        cabinet &	Stuff \\
        bed &	Thing \\
        chair &	Thing \\
        sofa &	Thing \\
        table &	Stuff \\
        door &	Stuff \\
        window &	Stuff \\
        counter &	Stuff \\
        shelves &	Stuff \\
        curtain &	Stuff \\
        ceiling &	Stuff \\
        refridgerator &	Thing \\
        television &	Thing \\
        person &	Thing \\
        toilet &	Thing \\
        sink &	Thing \\
        lamp &	Stuff \\
        bag &	Thing \\
        otherprop &	Stuff \\
    \bottomrule
    \end{tabular}
    \quad
    \begin{tabular}{l|l}
    \toprule
        Class & Type \\
    \midrule
        wall &	Stuff \\
floor &	Stuff \\
cabinet &	Stuff \\
bed &	Things \\
chair &	Things \\
sofa &	Things \\
table &	Stuff \\
door &	Stuff \\
window &	Stuff \\
counter &	Stuff \\
shelves &	Stuff \\
curtain &	Stuff \\
ceiling &	Stuff \\
refridgerator &	Things \\
television &	Things \\
person &	Things \\
toilet &	Things \\
sink &	Things \\
lamp &	Stuff \\
bag &	Things \\
bottle &	Things \\
cup &	Things \\
keyboard &	Things \\
mouse &	Things \\
book &	Things \\
laptop &	Things \\
blanket &	Stuff \\
pillow &	Things \\
clock &	Stuff \\
cellphone &	Things \\
otherprop &	Stuff \\

    \bottomrule
    \end{tabular}
    }
    
    \caption{Classes and their type (\textit{stuff} or \textit{thing}) for dataset experiments (left) and in the wild experiments (right).}
    \label{tab:supp_classes}
    \end{center}
\end{table}

\section{Additional Results}
\label{sec:supp_results}
Tab.~\ref{tab:supp_pq} reports the conventional PQ scores between our method and Mask2Former~\cite{cheng2022masked}. As mentioned in the main paper, this does not take into account the instance consistency across the scene, since matching between ground-truth and predicted instances is done on a per-frame basis. We further report SQ$\mathtt{scene}$ and RQ$\mathtt{scene}$ in Tab.~\ref{tab:supp_results}.

\end{appendix}

\end{document}